# Interpretable Models in ANNs

Yang Li

University of Victoria

leoli3737@gmail.com

## Abstract

Artificial neural networks are often very complex and too deep for a human to understand. As a result, they are usually referred to as black boxes. For a lot of real-world problems, the underlying pattern itself is very complicated, such that an analytic solution does not exist. However, in some cases, laws of physics, for example, the pattern can be described by relatively simple mathematical expressions. In that case, we want to get a readable equation rather than a black box. In this paper, we try to find a way to explain a network and extract a human-readable equation that describes the model.

# 1. Introduction

## 1.1. A General View

The term "artificial intelligence" (AI) refers to intelligence demonstrated by man-made machines, in contrast to the natural intelligence of living creatures such as we humans. From the view of computer science, it talks about the study of intelligent agents [1]: A device that is aware of the environment provided and takes actions that try to maximize its chance of successfully achieving some goals. It's often associated with the terms "learning" and "problem-solving".

Since the invention of the computer, many researchers have been exploring the field of AI from the view of computer science via various methods. The field of machine learning (ML) takes the approach of developing computer algorithms that improve through experience [2]. During the last decade, as one of the approaches in ML, artificial neural networks (ANN) have been becoming a more and more popular approach to a lot of real-world complicated problems such as image recognition, decision making, control systems and natural language processing based on complex mathematical models. AlphaGo Zero [3] developed by Google, for instance, is an AI that plays and beats human players in the game of Go. This is mainly due to the increasing computing power of our machines and the availability of a tremendous amount of data. The study of ANN can trace back to [4]. ANN is vaguely inspired by the biological neural networks in our human brain. It's believed that the network "learns" to perform a task by considering previously known examples and concluding patterns.

In modern ANN studies, quite a lot of achievements have been made in the mathematical theories behind the ML principles, and recently, we met some significant milestones in applied fields like self-driving vehicles and natural language processing. However, it is still very common that ANN models are referred to as black boxes, as the network is too complex for a human to interpret. Little effort in the studies of the interpretability of ANN models has been made. As a result, we humans cannot extract any knowledge from the network and do not have much control during the learning process. Although in some areas such as image recognition, the model itself might be fundamentally complex and thus it becomes impossible to describe it simply that a human brain can comprehend, in some other areas, on the other hand, the model is not so hard to describe. Such areas like physics usually have simple laws that can be described using relatively short mathematical expressions. In that case, we would like to find the expression instead of a black box, given that they both work accurately. The main purpose of this paper is to provide a methodology that helps to extract a highly readable mathematical expression from a given ANN that describes a relationship between different variables. Also, as by-products of that, two other problems are considered. First, as we will see later, ANN can only approximate functions on a bounded range of inputs. A model with better extendibility than the original ANN can be found, which should have better accuracy on inputs outside of the range in which the original ANN works. Second, a logical approach to developing the ANN which is different from learning empirical examples is introduced in this paper.

Nowadays, a lot of programming language libraries for ML have been developed by different companies and organizations. One of the most popular ones among them today is TensorFlow [5]. It has outstanding performance and many easy to use application programming interfaces (API), including Keras [6] which is the one used in the experiment part of this paper.

An ANN consists of a number (usually quite large) of neurons. A neuron will perform different tasks in various types of structures of the ANN. Each neuron has an activation function that mimics the process of a biological neuron being "activated". In this paper, we will study the fully connected feed-forward neural network with the rectified linear unit (ReLU) as the activation function.

For the rest of this paper, the abbreviations for fully connected feed-forward neural network, "ANN", "neural network", and "network" will be used interchangeably.

**1.2. Purpose**

The purpose of this paper is to study and examine ReLU networks to provide a high-level description of how they approximate functions and how to interpret them. In the methods section of this paper, we consider viable solutions to the following problems. The first problem is the main focus of this paper, and the latter two are by-products of a solution to the first problem.

- Given a ReLU network that approximates some target function on a bounded range of inputs, find a mathematical expression that approximates the original target function.

- Based on the previous problem, the expression should have better extendibility than the ReLU network, meaning that the expression can be applied to a wider range of inputs with better accuracy.

- Given a mathematical expression, build a ReLU network that approximates it on some bounded range of inputs without training.

**1.3. Outline**

In section 2, we provide the background knowledge of ANN in terms of 3 aspects. How it is formally defined and perform computations internally; how it is trained and learn from experience using a method called backpropagation; and lastly, how it universally approximates a target function while using ReLU as activation functions. Section 3 focuses on a detailed description of the approaches used in this study that tries to address the three problems mentioned above. Section 4 presents the results of the numerical experiments conducted using the methods demonstrated in section 3. Section 5 discusses the results of the previous section. A conclusion is drawn at the end of this section, followed by acknowledgments, the reference list, and the appendix.

## 2. Background

**2.1. What Is ANN**

A network is formally defined as the following. An ANN is a mathematical function denoted $F_A$ that takes as input a real-valued vector and computes as output a real-valued vector. The ANN consists of at least two layers sequentially, where the first layer is called the input layer, the last layer is called the output layer, and a layer in the middle, if there is any, is called a hidden layer. Each layer consists of many neurons that perform some computation. In the input layer, the number of neurons is the size of the input vector, with each neuron holds the value of one entry of the vector. The same thing happens for the output layer. Starting from the second layer, each neuron takes as input all the output values of neurons from the last layer and then computes a weighted sum of those inputs. This sum is to be added with a bias term and then sent into an activation function. Finally, the output of the activation function will be the output of this neuron. Each layer has its activation function and each neuron has its weights and bias. Using the matrix and vector notation (the arrow notation will be omitted for the rest of this paper when it's clear from context for conciseness), a one hidden layer ANN can be expressed as follows. Let $\vec{x}$ be the input vector. Let $W_1$ and $W_2$ be the matrix representing the weights of the hidden layer and output layer, respectively, where each row is the weights for one neuron. Let $\vec{b_1}$ and $\vec{b_2}$ be the vector representing the biases for those two layers, respectively, where each entry is the bias for one neuron. Finally, let $\sigma_1$ and $\sigma_2$ be activation functions.

$$F_A(\vec{x}) = \sigma_2\left(W_2 \sigma_1\left(W_1 \vec{x} + \vec{b_1}\right) + \vec{b_2}\right) \tag{1}$$

Depending on how the activation functions are defined, the network can perform different tasks. For example, if the task is to compute some kind of probability, $\sigma_2$ can be the sigmoid function $\sigma(x) = \frac{1}{1+e^{-x}}$, as the output is in the range $(0,1)$. If the task is regression, $\sigma_2$ can be the identity function. For this paper, we consider only regression problems (function approximation). ReLU has become a popular alternative to the sigmoid function recently, due to its computational simplicity and other nice properties [7]. The ReLU function is formally defined as

$$R: \mathbb{R} \to \mathbb{R}, \qquad R(x) = \max(\{0, x\}) = \begin{cases} 0, & if\ x \leq 0 \\ x, & if\ x > 0 \end{cases} \tag{2}$$

We will focus on this specific activation function for the latter parts.

**2.2. Backpropagation**

Because the detailed training process of the neural network is not part of the main focus of this paper and will not be mentioned again for the rest of this paper, here we only provide a high-level view of how the training is performed. The network learns the model using a method called backpropagation [8]. Initially, the weights and biases in an ANN are randomly assigned. To train the network is to provide examples and try to adjust the weights and biases in the network such that the new output of the ANN is closer to the target value according to the example. Given an example input-output pair, we measure how close the computed output and the target output is through something called the loss function, also referred to as the error function. For example, a commonly used loss function is the squared loss. Let $y$ be the true vector value according to the example, and let $\hat{y}$ be the predicted vector by the ANN. It is formally defined as

$$L(y, \hat{y}) = (y - \hat{y})^2 \tag{3}$$

We then use the gradient descent method (originally proposed by A. L. Cauchy in 1847) to minimize the loss function. Thus, the network is refined and will hopefully generate a better output that is closer to the true value next time. The high-level idea is that if we consider the function of the error measured on that specific example in terms of the weights and biases, there is a direction in which the error is dropping most rapidly according to the gradient. We adjust the weights and biases in that direction, so the error is reduced on that specific example. Note that this is essentially a mathematical optimization problem. It is also possible (normally better) to consider the average error measured among multiple examples [9]. Such a set of examples is called a mini-batch. As we train the network on a set of examples called the training sample set, the empirical error will decline. In the end, we are expected to get a good ANN that grasps some kind of pattern between the input and output.

**2.3. A View of Function Approximation**

The universal approximation theorem [10] states that on any compact subset of $\mathbb{R}^n$, and for any function that is continuous in that set, there is a feed-forward network with a single hidden layer that can approximate the function as close as we want. under mild assumptions on the activation function. The universal approximation theorem for width-unbounded networks can be expressed as follows:

For all compact subset $C \subsetneq \mathbb{R}^n$, for all continuous function $f: C \to \mathbb{R}$, for all $\varepsilon > 0$, there exists one hidden layer fully connected network $A$ with some activation function, such that the function $F_A: \mathbb{R}^n \to \mathbb{R}$ represented by this network satisfies

$$|F_A(x) - f(x)| < \varepsilon \tag{4}$$

[11] proved that this holds for networks that use ReLU as activation functions. It's been shown that a feed-forward artificial neural network is essentially a linear spline [12, 13]. We provide a high-level explanation in this paper. Say you have a network using ReLU as activation functions. The network $A$ that takes as input $x \in \mathbb{R}^n$, an n-dimensional real-valued vector, and computes as output $y = F_A(x) \in \mathbb{R}$, a real number. Recall equation (2), where ReLU is defined as $\max(\{0, x\})$. We say the neuron is activated when the output is greater than 0. We can see when the ReLU is activated, it is simply the identity function $f(x) = x$. In other words, it describes a straight line on the graph. This is the key idea of how a ReLU network approximate functions.

Consider a specific input vector $x$, let the network compute y as output, then we remove all the neurons in the hidden layers that are not activated. When a neuron is not activated it outputs 0, thus it contributes nothing to the next layer, no matter what its outgoing weights are, so the output of the network is still y. Now we remove the activation functions, the remaining network still outputs y since ReLU is just the identity function when it is activated and removing it changes nothing. Now we can see y is computed as an affine transformation of its last layer's outputs, and so is every neuron in its last layer. Since an affine transformation of another affine transformation of a set of real values $S$ is still just an affine transformation of values in the set $S$, ultimately the output of the neural network is an affine transformation of the entries of the input $x$, denoted with $w \cdot x + b$. As long as the set of activated neurons remains the same, the network will keep computing as output $F_A(x) = w \cdot x + b$.

As we change the input $x$, some new neurons might be activated. We know the process of changing the set of activated neurons is discrete, thus there must be a range of input, such that the original network will keep computing as output $F_A(\tilde{x}) = w \cdot \tilde{x} + b$, for all $\tilde{x}$ that is in that range (i.e. close enough to $x$). The idea is that the network is just using a local linear approximation to predict the target function near $x$. Thus, it is essentially a linear spline. More formally, we have,

For all $x \in \mathbb{R}^n$, there exists a $w \in \mathbb{R}^n$ and a compact subset $C$ where $\{x\} \subsetneq C \subsetneq \mathbb{R}^n$, such that for all $\tilde{x} \in C$,

$$F_A(\tilde{x}) = w \cdot \tilde{x} + b \tag{5}$$

This tells us a ReLU network is drawing the target function using tiny flat pieces. The accuracy, of course, depends on the number of knots and the distances between them (i.e. step size). This explains why ReLU networks can approximate any function that is continuous in the compact subset where we are making our approximation. As the function is continuous in the compact subset, the change in the function's output can get arbitrarily small if the change in the input is small. More formally, a function is continuous in an input domain $D$ means

$$\forall\, x_0 \in D, \forall\, \varepsilon > 0, \exists\, \delta > 0, \forall\, x \in D:$$

$$|x - x_0| < \delta \Rightarrow |f(x) - f(x_0)| < \varepsilon \tag{6}$$

Thus, using small enough linear pieces, the error can be arbitrarily small for any continuous target function.

## 3. Methods

### 3.1. From Slope to Derivative

Now we know how ReLU networks approximate functions. We then use this property of ReLU to construct an expression that approximates the target function. For simplicity, let's consider a simple network $A$ (not necessarily using ReLU) where the function represented by it is $F_A: \mathbb{R} \to \mathbb{R}$ which approximates the target function $f$ on some closed interval $[a, b]$. Assume $f$ satisfies condition (6) on $[a, b]$ and thus it is said to be continuous on $[a, b]$ (i.e. $f$ is not ill-conditioned). If $F_A$ satisfies inequality (4) with a negligible $\varepsilon$, we say that $F_A(x) \approx f(x)$ on $[a, b]$ with error bound $\varepsilon$. Although $f$ might not be an analytic function and thus getting a single mathematical expression of it is not possible, we can still define an analytic function that approximates $f$ by building a refinement of $F_A$, denoted $\hat{f}$. If on the interval $[a, b]$, $F_A(x) \approx f(x)$ with error bound $\varepsilon$, and $\hat{f}(x) \approx F_A(x)$ with error bound $\delta$, then obviously $\hat{f}(x) \approx f(x)$ with error bound $\varepsilon + \delta$. The total error bound can be arbitrarily small if and only if $\delta$ can be arbitrarily small. Later we will see how to achieve this goal.

A nice property about the linear spline $F_A$ is that the slope of the line segment is the derivative of $F_A$ which is easily computed. How can derivative help us with interpreting the model represented by an ANN? Let $f^{(k)}$ be the $k$-th derivative of $f$, so $f^{(0)} = f$. Consider the Taylor's theorem (originally proposed by B. Taylor in 1712):

> Let $f: \mathbb{R} \to \mathbb{R}$ be an $n$ times differentiable function at the point $x_0 \in \mathbb{R}$, then there exists an $h_n: \mathbb{R} \to \mathbb{R}$ called the Peano remainder, where $\lim_{x \to x_0} h_n(x) = 0$, such that

$$f(x) = \sum_{k=0}^{n} \frac{(x - x_0)^k}{k!} f^{(k)}(x_0) + (x - x_0)^k h_n(x) \tag{7}$$

> Furthermore, if $f$ is real analytic, then locally

$$f(x) = \sum_{k=0}^{\infty} \frac{(x - x_0)^k}{k!} f^{(k)}(x_0) \tag{8}$$

Locally means that the equality holds on some interval $(a, b)$ that includes $x_0$ if there is no discontinuity on $(a, b)$. Now we show how to build the function $\hat{f}: [a, b] \to \mathbb{R}$, which tries to recover $f$ by using a smooth approximation of $F_A$. This is easy for ANNs that use ReLU as activation functions, which are essentially linear splines. There are infinitely many functions that can have the linear spline $F_A$, so let's pick one that is easy to work with, namely a real analytic function $\hat{f}$. The final expression that approximates the model will be stated as a finite number of terms of its Taylor series.

To get a Taylor series of $\hat{f}$, we need to get the derivatives of it. Once we have different orders of its derivatives, we can just pick a point $x_0$ and plug values in to get the Taylor expansion of the function at $x_0$. Consider one of the line segments of the linear spline, denoted $l(x) = kx + c$ on the domain $[m, n]$. By the nature of splines, the knots, $a$ and $b$ are accurate values of $\hat{f}$. The average derivative of $\hat{f}$ over $[m, n]$ is, by the fundamental theorem of calculus, defined as

$$\frac{1}{n-m} \int_m^n \frac{d\hat{f}(x)}{dx} dx$$

$$= \frac{\hat{f}(n) - \hat{f}(m)}{n-m} = \frac{l(n) - l(m)}{n-m} = \frac{kn + c - (km + c)}{n-m} = \frac{k(n-m)}{n-m} = k \quad (9)$$

The average derivative of $\hat{f}$ over $[m, n]$ is simply the slope of the line segment defined by $F_A$ on $[m, n]$. $F_A$ is almost everywhere differentiable, except at the knots. $F_A'(x)$ is also the slope of the line segment at $x$. Recall that $\hat{f} \approx F_A(x)$ with error bound $\delta$. If $\delta$ is small enough, we have $F_A'(x) \approx \hat{f}'(x)$ on $[a, b]$. Note this property holds because of the nice property of linear spline. It may not hold for activation functions other than ReLU. In other words, the $F_A'$ is the slope which is also the average values of $\hat{f}'$ which can be a good approximation to $\hat{f}'$ itself. The slope is easily computed if we look at the weights inside the neural network. Let the output of an activated neuron $N$ be $N(\vec{h}) = \vec{w} \cdot \vec{h} + b$ where $\vec{w}$ and $b$ are the weights and bias for that neuron and $\vec{h}$ is the vector of outputs from its last layer and let $\vec{h}_i$ denotes its $i$-th entry. The deactivated neurons just output 0. To compute $F_A'(x)$ is to compute $\frac{dy(x)}{dx}$ where $y$ is the neuron in the output layer. The derivative of the output of a neuron $N$ can be computed recursively using the chain rule.

$$\frac{dN(x)}{dx} = \vec{w} \cdot \langle \frac{d\vec{h}_1}{dx}, \dots, \frac{d\vec{h}_n}{dx} \rangle \quad (10)$$

Since $\hat{f}$ is real analytic, $\hat{f}'$ is real analytic, too. Thus, there is another ANN learns a good approximation to it. To get training examples for our new ANN that will learn $\hat{f}'$, we apply this method of computing slopes to different line segments of the linear spline. Then, we get a table of discrete approximate values of $\hat{f}'(x)$ on $[a, b]$, which can be used as the training sample set for the new ANN. We can obtain as many samples as we need, so the new ANN can fit the derivative of the previous ANN arbitrarily well. Then, we can repeat this process to get an approximation to $\hat{f}'', \hat{f}''',$ and so on. Once we have those, all we need to do is to pick a value $x_0$ and plug it into the neural networks to get different orders of derivatives. Let $F_{A^{(k)}}$ be the function represented by the ANN that learns $\hat{f}^{(k)}$, so $F_{A^{(0)}} = F_A$ learns $\hat{f}$. Plug those values into (8). We get

$$f(x) \approx F_A(x) \approx \hat{f}(x) = \sum_{k=0}^{\infty} \frac{(x-x_0)^k}{k!} \hat{f}^{(k)}(x_0) \approx \sum_{k=0}^{\infty} \frac{(x-x_0)^k}{k!} F_{A^{(k)}}(x_0) \quad (11)$$

Theoretically, if the neural networks are accurate, the infinite series should be the same no matter which point we chose as our $x_0$, but with some inevitable error in the networks, it is a good idea that we pick several $x_0$, get a Taylor series from each of them, and take the average. The average error must be no greater than the greatest error among them, so it has better accuracy. Let $T_i$ be the Taylor series obtained from the $i$-th choice of $x_0$. We use the following average as our Taylor series.

$$\hat{f}(x) \approx \frac{1}{n} \sum_{i=1}^{n} T_i(x) \quad (12)$$

Combining expression (11) with expression (12) we have

$$\hat{f}(x) \approx \frac{1}{n} \sum_{i=1}^{n} \sum_{k=0}^{\infty} \frac{(x-x_{0_i})^k}{k!} F_{A^{(k)}}(x_{0_i}) \quad (13)$$

Practically speaking, we have to stop the algorithm at some point. If we are lucky, in the case of the target function is a polynomial, we get a finite series and the algorithm stops as we get a constant derivative of 0. More commonly, however, if the target function is not a simple polynomial, we end up with an infinite Taylor series. Based on the precision we desire depending on the practical situation of the real-world problem, we need to chop off the series at some point and obtain a finite polynomial. As we chop off the series at some point, the series is only a local approximation centred at our choice of $x_0$. In that case, take an average of several Taylor series can lead to a worse result, because each of them is for a different $x_0$. We should just choose the point in the range where we want to approximate. Let $m$ be the order of Taylor series we want. We compute the following as our final approximation to the refined version of $F_A(x)$, which goes back to approximating the original target function $f$.

$$f(x) \approx F_A(x) \approx \hat{f}(x) \approx \sum_{k=0}^{m} \frac{(x-x_{0_i})^k}{k!} F_{A^{(k)}}(x_{0_i}) \quad (14)$$

Note that it is also possible to replace $F_{A^{(k)}}(x_{0_i})$ with $F'_{A^{(k-1)}}(x_{0_i})$, but the result is not as good. The details will be explained in section 3.4.

The algorithm is demonstrated in a high-level sense in the following pseudocode. Note that "//" is followed by a line of comments.

1D Taylor series

Input: A set of training examples $S = \{(x_1, f(x_1)), \ldots, (x_j, f(x_j))\}$, and a natural number $m$ which represents the order of desired Taylor series

Output: A polynomial $P$ of order $m$ that approximates $f$

Algorithm:

  // get the smallest closed interval that encloses the inputs in $S$ for future references

  $[a, b] \leftarrow [\min x_i, \max x_i]$

  $k \leftarrow 0$

  while $k \leq m$ do:

    $F_{A^{(k)}} \leftarrow$ a ReLU network that has been trained on the sample set $S$

    // obtain a new training set from $[a, b]$ where $t$ the predefined training set size

    $X \leftarrow \{\hat{x}_1, \ldots, \hat{x}_t\}$ where each $\hat{x}_i \in [a, b]$

    $S \leftarrow \{(\hat{x}_1, F'_{A^{(k)}}(\hat{x}_1)), \ldots, (\hat{x}_t, F'_{A^{(k)}}(\hat{x}_t))\}$

    $k \leftarrow k + 1$

  end

  if $F'_{A^{(k-1)}}(\hat{x}_i) \neq 0$ for any $\hat{x}_i$ then

    $x_0 \leftarrow$ some value in $[a, b]$

    return $\sum_{k=0}^{m} \frac{(x-x_0)^k}{k!} F_{A^{(k)}}(x_0)$

  end

  // take an average among $n$ polynomials where $n$ is a predefined positive integer

  $X \leftarrow \{x_{0_1}, \ldots, x_{0_n}\}$ where each $x_{0_i} \in [a, b]$

  $i \leftarrow 1$

  while $i \leq n$ do:

    $P_i(x) \leftarrow \sum_{k=0}^{m} \frac{(x-x_{0_i})^k}{k!} F_{A^{(k)}}(x_{0_i})$

    $i \leftarrow i + 1$

  end

  return $\frac{1}{n} \sum_{i=1}^{n} P_i(x)$

### 3.2. From Derivative to Gradient

Using the same idea, one can easily adapt the method to higher dimensions. For convenience, let's denote the functions with multiple inputs with functions that take a vector as input. The arrow notation is used to avoid the confusion in subscripts. Let the function $f: \mathbb{R}^n \to \mathbb{R}$ be an analytic function, $\vec{x}$ be the input vector and $\vec{x_0}$ be a specific input. Let the notation $\vec{x}_i$ denote the $i$-th entry of $\vec{x}$. Consider the Taylor series in $n$-dimensional space:

$$f(\vec{x}) = \sum_{k_1=0}^{\infty} \cdots \sum_{k_n=0}^{\infty} \frac{\prod_{i=1}^{n}(\vec{x}_i - \vec{x_0}_i)^{k_i}}{\prod_{i=1}^{n} k_i!} \frac{\partial^{\sum_{i=1}^{n} k_i} f}{\prod_{i=1}^{n} \partial \vec{x}_i^{k_i}}(\vec{x_0}) \quad (15)$$

Applying the multi-index notation, the series can be written as a more compact form,

$$f(\vec{x}) = \sum_{|\alpha| \geq 0} \frac{(\vec{x} - \vec{x_0})^{\alpha}}{\alpha!} \partial^{\alpha} f(\vec{x_0}) \quad (16)$$

Here is an example of a second-order Taylor series $P$ of a function $f$ with 2-dimensional input. Let $\nabla^2 f$ be the Hessian matrix notation that represents second-order partial derivatives.

$$P(\vec{x}) = f(\vec{x_0}) + (\vec{x} - \vec{x_0}) \cdot \nabla f(\vec{x_0}) + \frac{(\vec{x} - \vec{x_0})^T}{2!} \cdot \nabla^2 f(\vec{x_0}) \cdot (\vec{x} - \vec{x_0}) \quad (17)$$

The partial derivatives are also easily computed looking at the weights inside the ReLU network, as the function is made of tiny flat pieces. Equation (10) is now adapted to the following form

$$\frac{\partial N(\vec{x})}{\partial \vec{x}_i} = \vec{w} \cdot \langle \frac{\partial \vec{h}_1}{\partial \vec{x}_i}, \ldots, \frac{\partial \vec{h}_n}{\partial \vec{x}_i} \rangle \quad (18)$$

Based on the algorithm for 1D Taylor series in section 3.1, the general algorithm is described in a higher-level language in the following pseudocode. It is similar to the one illustrated in section 3.1, except that here it computes partial derivatives needed for the Taylor series instead of derivatives. The interval we were working on, $[a, b]$, is expanded to a compact subset of $\mathbb{R}^n$ naturally. The sequence of ReLU networks is trained in a breadth-first search (BFS) fashion, where the $f$ is the root node with children $\frac{\partial f}{\partial \vec{x}_1}, \ldots, \frac{\partial f}{\partial \vec{x}_n}$. A nice property of partial derivatives and save us some effort. Since the order in which we take partial derivatives does not matter for analytic functions (e.g. $f_{xy} = f_{yx}$), we only need to compute some branches.

> General Taylor series
>
> Input: A set of training examples $S = \{(\vec{x_1}, f(\vec{x_1})), \ldots, (\vec{x_j}, f(\vec{x_j}))\}$, and a natural number $m$ which represents the order of desired Taylor series
>
> Output: A polynomial $P$ of order $m$ that approximates $f$
>
> Algorithm:
>
>     $C \leftarrow$ the smallest hyperrectangle that encloses all $\vec{x_i}$
>
>     // $\alpha$ is a multi-index
>
>     $R \leftarrow$ a set of ReLU networks trained for each of $\bigcup_{|\alpha| \leq m} \partial^\alpha f$ on inputs in $C$
>
>     if $\partial^\alpha f(\vec{x}) \neq 0$ for any $\alpha$ where $|\alpha| = m$ then
>
>         $x_0 \leftarrow$ some value in $C$
>
>         return $\sum_{|\alpha| \geq 0} \frac{(\vec{x} - \vec{x_0})^\alpha}{\alpha!} \partial^\alpha f(\vec{x_0})$
>
>     end
>
>     // take an average among $n$ polynomials where $n$ is a predefined positive integer
>
>     $X \leftarrow \{\vec{x_{0_1}}, \ldots, \vec{x_{0_n}}\}$ where each $\vec{x_{0_i}} \in C$
>
>     $i \leftarrow 1$
>
>     while $i \leq n$ do:
>
>         $P_i(\vec{x}) \leftarrow \sum_{|\alpha| \geq 0} \frac{(\vec{x} - \vec{x_{0_i}})^\alpha}{\alpha!} \partial^\alpha f(\vec{x_{0_i}})$
>
>         $i \leftarrow i + 1$
>
>     end
>
>     return $\frac{1}{n} \sum_{i=1}^{n} P_i(\vec{x})$

Although it was not the main goal of this study that this method, as a side product, provides a possible way of obtaining a model with better extendibility than the original ANN. As the universal approximation theorem described in inequality (4), the ANN is only expected to approximate the target function on a closed interval. Any ReLU network cannot realize the simple polynomial $f(x) = x^2$, as the linear spline eventually only grows linearly. With the method introduced in this section, there is a chance to recover the target function on a universal scale or at least get better accuracy, if the target function is a finite polynomial. Later we will see similar examples in the experiments section.

### 3.3. Determine the Order

One problem remains. How do we select the order of Taylor series we want? Unfortunately, this is a hopeless problem, unless we get unlimited oracle access (in the sense of a black box) to the target function. Because we only have sample data in a limited range, no numerical analysis method can tell the difference between some functions without perfect precision.

This is exactly what happened hundreds of years ago when Galileo Galilei (oddly enough, most people only know his first name) came up with an equation for velocity adding, $v = v_1 + v_2$. However, it is not until centuries later that A. Einstein proved that is wrong according to his special theory of relativity. They are two of the greatest scientists of all time, so what went wrong? According to the special theory of relativity, a better (maybe correct) model for velocity along one axis is $v = \frac{v_1 + v_2}{1 + v_1 v_2 / c^2}$ where $c \approx 3 \times 10^8$ meters per second (the speed of light). We can see that two models are very close under low speed as the denominator $1 + v_1 v_2 / c^2 \approx 1$. The two are identical in cases where we use a measuring device or a computer with imperfect precision. If we cannot do experiments under very high speed, we will never realize the problem.

However, there is still something we can do. Suppose we have an ANN that learned the target function, then we can use the following trick. Fix all input variables except for one of them. Let's call it $x$. Now let the network compute different outputs for different values of $x$. We then obtain a graph of the target function's 2D projection, denoted $f(x)$. Many numerical analysis methods can be applied. We could, for example, use the finite-difference method (first proposed by L. Euler in 1768) to guess the degree of $x$ in the polynomial. The whole idea of the Taylor approximation in section 3.1 and 3.2 is a finite-difference method. Or if we get access to very big values of $x$, then we can draw the graph of $\frac{\log f(x)}{\log x}$. One can easily show that if $f(x)$ is a polynomial of $x$ where the coefficient of the term with the highest degree is positive, then

$$\lim_{x \to \infty} \frac{\log f(x)}{\log x} = c \tag{19}$$

where $c$ is the degree of $f(x)$. If the coefficient of the term with the highest degree is negative and the function seems to decrease as $x \to \infty$, just try $\frac{\log -f(x)}{\log x}$ instead. If there are negative values of $f(x)$, we can use $\frac{\log[f(x)+k]}{\log x}$ to lift the function. This will also work the same way. We do that for each variable to get the overall degree of the polynomial. If we are lucky, the underlying model is indeed a polynomial then we might get the correct degree. Otherwise, choosing a high enough order is just a matter of accuracy. What do statisticians say? "All models are wrong, but some are useful".

### 3.4. Technical Notes

During the experiments, it has been found that the predicted values near the boundary of the training data set, unsurprisingly, have lower accuracy. A useful trick for training the ANNs is to sample the derivatives in a smaller area than where the previous training sample set was from. Thus, we avoid the bad training values of derivative on the edges. Another noticeable point is that when the underlying model is given by a finite polynomial, the Taylor series will come to an end when it gets derivatives of constant 0. We should choose the precision based on training error (empirical loss). For example, if the training error is 0.01, then it is meaningless to compute the derivatives to the third decimal place. This helps a lot when it comes to terminating the Taylor series when it is given by a finite polynomial as the ending derivatives will be very close to 0.

As mentioned after equation (14). We can use either the calculated slope or the learned derivative as the training sample. During the experiments, it's found that the learned derivative might be a better choice. ReLU networks are good at fitting a target function, but the derivatives are not so accurate. In the figure below, the derivative (slopes) of the ANN is shown in blue, and the target function's actual derivative is shown in orange.

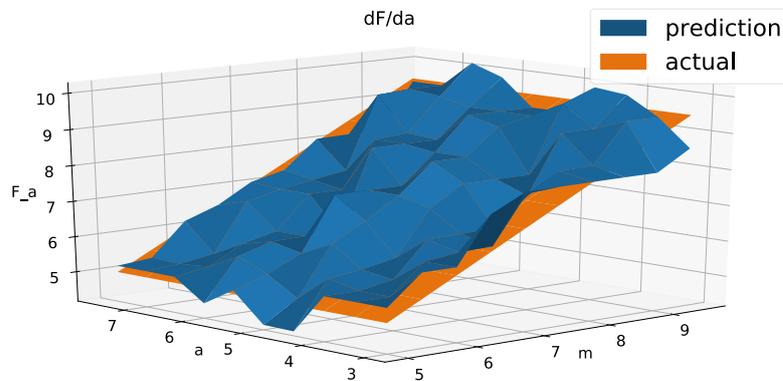

**Figure 3.4.1**

We can see that the overall trend is not bad, but not the individual values. Letting another ANN to learn it with some regularization (limiting the model complexity) can address this problem.

### 3.5. Reverse Engineering

Now we know how to extract an expression from a ReLU network. Theoretically although very tediously, we can do the reverse. Given a mathematical expression as the continuous target function, we can manually set weights in the ReLU network to make it fit. It is shown in section 2.3 that ReLU networks are line splines. To make the line spline of a given function, we simply put little linear pieces according to the function's derivative. The idea is to have neurons that check the range of input, and a neuron for a linear summation over different dimensions. Overall, the network has two hidden layers. This is a similar idea to [14] where the sigmoid function as the activation function is considered, but the derivatives do not match. In this section, we provide a method that not only matches the function values, but also the derivatives.

For example, if we want to draw the function $f(x) = x^2$, we first calculate its derivative, namely $2x$. Now, say we want to use three line segments on the interval $(1,2], (2,3], (3,4]$, respectively. Using the middle value of each interval to calculate the derivative, we need lines of slope 3, 5, 7, respectively. We build neurons in the first hidden layer that work in triples to perform the task of identifying the range of inputs and build a neuron in the second hidden layer to sum the triple up. For convenience, let's define $-\infty$ to be negative infinity (or just a negative number that's small enough to work). Define $(-\infty)0 = 0$. Consider the incoming weights for 3 neurons in the first hidden layer. Set the weights to be –1, 1 and 1, respectively. Set the biases to be 1, –1 and –2, respectively. Now we can see that, the first neuron will be activated when $x < 1$; the second one will be activated when $x > 1$; and the last one will be activated when $x > 2$. Set the outgoing weights of them to be $-\infty$, 3 and $-\infty$, respectively. When $x \in (1,2]$, only the second one is activated, and the triple together contribute $3(x - 1)$ to the overall output. When $x < 1$ or $x > 2$, however, one of the neurons with outgoing weight $-\infty$ is turned on and dominates the sum of the triple, so the triple together outputs a sum of $-\infty$. Thus, the summing neuron in the next layer will output 0 after ReLU. As a result, this unit of four neurons will put a line segment of slope 3 on the interval [1,2]. The next group of four is constructed for the next interval the same way. Then we successfully obtain a line spline based on derivatives. The last thing to do is to set the bias for the neurons in the second layer. This will have the effect of moving the line segment up and down. We just put it wherever fits the best.

Note that this method is essentially a constructive version of the universal approximation theorem for two hidden layers ReLU networks. One can easily generalize this method to higher dimensions. The triples will be updated to sets of triples that control each dimension. Note that the $-\infty$ weights and the summing neuron in the second layer together work essentially as a logical AND gate. As long as at least one of the inputs is no in the desired range the overall output will be 0. Also, if we accept the definition of $-\infty$, then the ReLU network will have the potential to fit piecewise functions (but not general discontinuous functions), because the ANN is now discontinuous.

### 3.6. A More General Sense

It is possible to solve (but not practically) any finite problem with the Taylor series. Let's formally define what a problem is and what it means to solve a problem. A problem is a function $p: Q \to A$, with question space $Q$ and answer space $A$. A solution to the problem $P$ is a function $s: Q \to A$ such that $\forall\, q \in Q: s(q) = p(q)$. Solving the problem means finding a computable solution. A problem is called finite when $|Q| < \infty, |A| < \infty$. Consider a finite problem $p: Q \to A$, let $|Q| = m$ and $|A| = n$. We can encode every input and output values using real numbers, as they are finite. We then draw the mapping from $m$ real values to their corresponding real values. That gives us a set of points $\{(q_1, P(q_1)), \dots, (q_m, P(q_m))\}$. By the interpolation theorem (as a result of the fundamental theorem of algebra), there is a unique polynomial of degree $m - 1$ that passes through all the points, denoted $S'$. Thus, the function $S: Q \to A, T(q) = DEC\left(S'\bigl(ENC(q)\bigr)\right)$ where $ENC$ and $DEC$ are the encoding and decoding for questions and answers, respectively, is a solution to the problem. $S'$ is a polynomial thus it is its own Taylor series.

## 4. Experiments

In each of the sections below, a target function is to be recovered using the algorithm introduced in section 3. Given that I do not have a powerful computing device, some simple functions are tested. All the underlying models tested are the ones that are impossible for ReLU networks to universally realize. Since one-hidden-layer ANNs already are universal function approximators, I am using only ReLU networks with a single hidden layer. It is easier to compute the gradient. For all the figures, blue represents the output values of the ANN and orange is the desired true value. The gap between the predicted values and the actual ones might appear larger than actual. Notice the scale of the axis.

### 4.1. A Simple Function

$$f(x) = 0.5x^2 - 2x + 1$$

To make the task harder, the training sample data points are not evenly spaced.

Results:

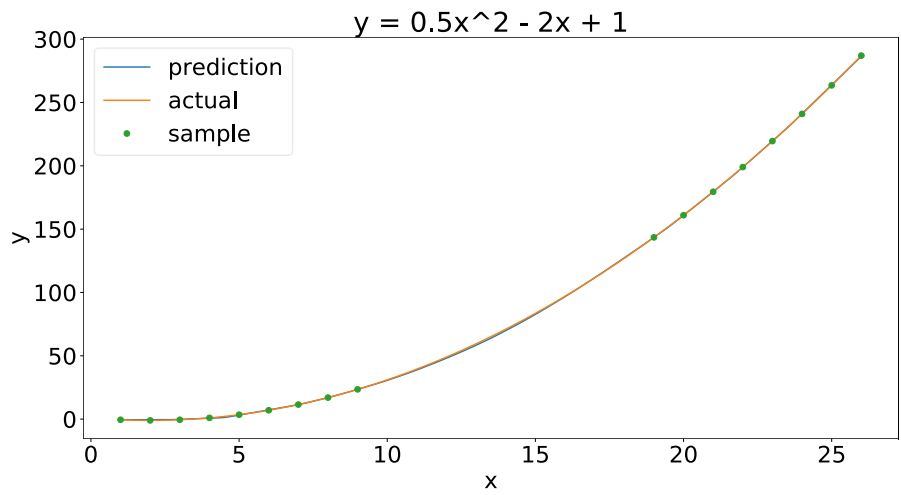

Figure 4.1.1

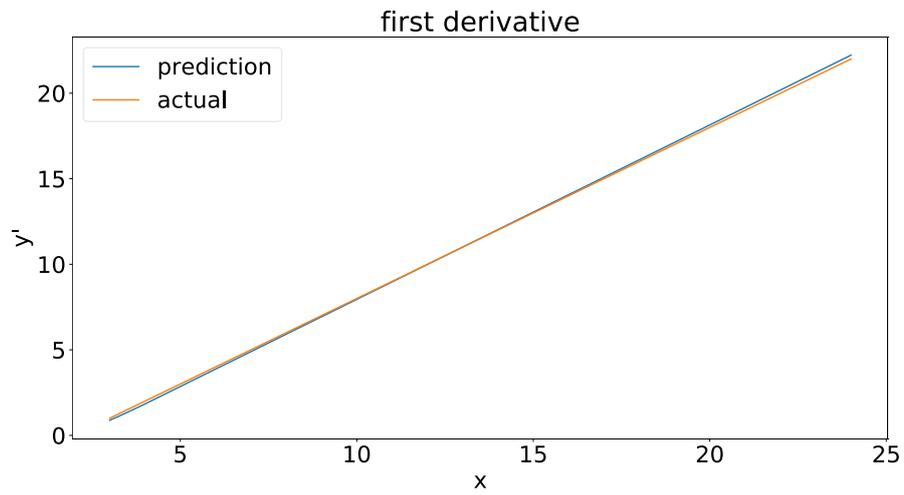

Figure 4.1.2

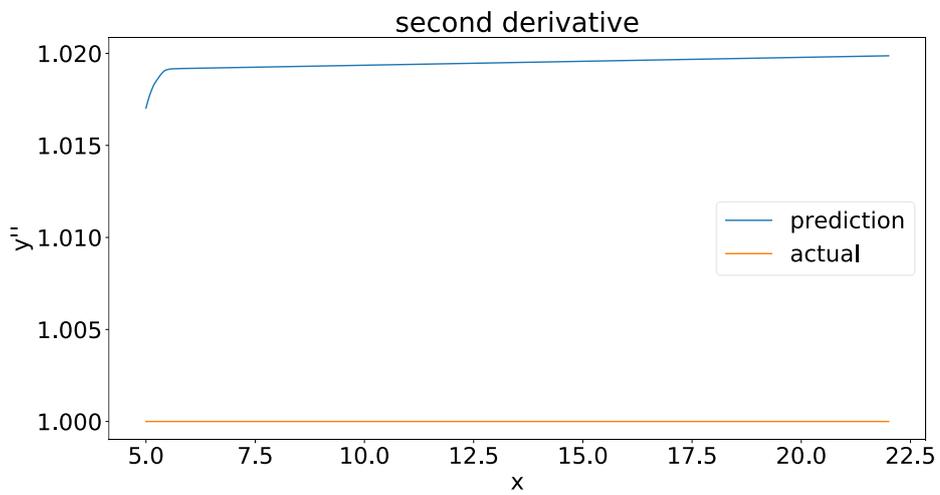

Figure 4.1.3

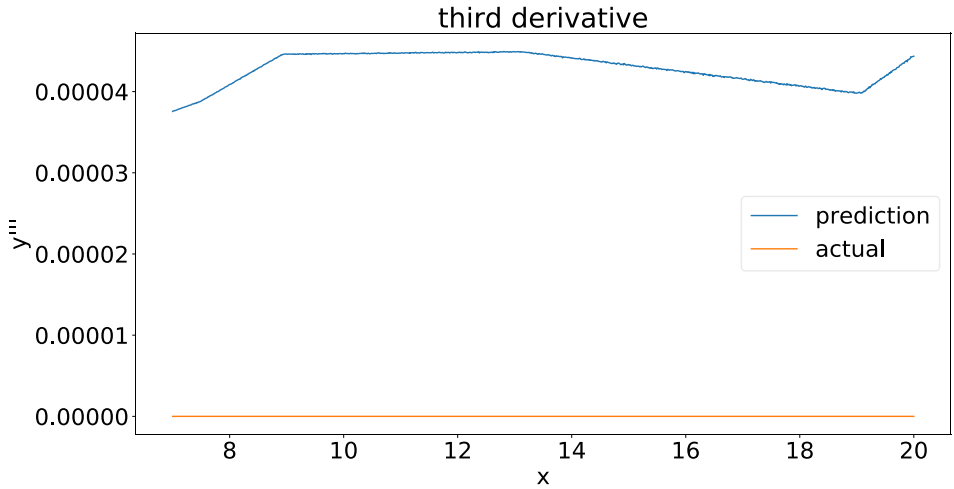

**Figure 4.1.4**

| $F(x_0)$ $x_0$ | $F_{A^{(0)}}(x_0)$ | $F_{A^{(1)}}(x_0)$ | $F_{A^{(2)}}(x_0)$ | $F_{A^{(3)}}(x_0)$ |
|---|---|---|---|---|
| 7 | 11.5 | 4.89 | 1.02 | 0.00 |
| 8 | 17.0 | 5.91 | 1.02 | 0.00 |
| 9 | 23.5 | 6.93 | 1.02 | 0.00 |
| 10 | 30.6 | 7.95 | 1.02 | 0.00 |
| 11 | 38.8 | 8.97 | 1.02 | 0.00 |

$$P_1(x) = \frac{(x-7)^0}{0!} 11.5 + \frac{(x-7)^1}{1!} 4.89 + \frac{(x-7)^2}{2!} 1.02 + \frac{(x-7)^3}{3!} (0.00)$$

$$= 0.51x^2 - 2.25x + 2.26$$

$$P_2(x) = \frac{(x-8)^0}{0!} 17.0 + \frac{(x-8)^1}{1!} 5.91 + \frac{(x-8)^2}{2!} 1.02 + \frac{(x-8)^3}{3!} (0.00)$$

$$= 0.51x^2 - 2.25x + 2.36$$

$$P_3(x) = \frac{(x-9)^0}{0!} 23.5 + \frac{(x-9)^1}{1!} 6.93 + \frac{(x-9)^2}{2!} 1.02 + \frac{(x-9)^3}{3!} (0.00)$$

$$= 0.51x^2 - 2.25x + 2.44$$

$$P_4(x) = \frac{(x-10)^0}{0!} 30.6 + \frac{(x-10)^1}{1!} 7.95 + \frac{(x-10)^2}{2!} 1.02 + \frac{(x-10)^3}{3!} (0.00)$$

$$= 0.51x^2 - 2.25x + 2.1$$

$$P_5(x) = \frac{(x-11)^0}{0!} 38.8 + \frac{(x-11)^1}{1!} 8.97 + \frac{(x-11)^2}{2!} 1.02 + \frac{(x-11)^3}{3!}(0.00)$$
$$= 0.51x^2 - 2.25x + 1.84$$

$$\hat{f}(x) \approx \frac{1}{5}\sum_{i=1}^{5} P_i(x) = 0.51x^2 - 2.25x + 2.2$$

The absolute error upper bound of $\hat{f}(x)$ on the sample interval [1,26]: 1.46

The average error of $\hat{f}(x)$ on the sample interval: 0.401

The average error of the original ReLU network on the sample interval: 0.354

### 4.2. Newton's Second Law

$$F(m, a) = ma$$

To simulate actual physics experiments, a 0.1% measuring error is introduced. Real-world measuring tools should have accuracy way higher than 99.9%.

Results:

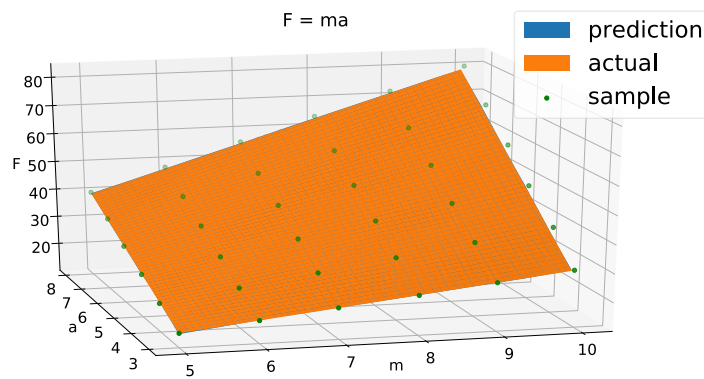

**Figure 4.2.1**

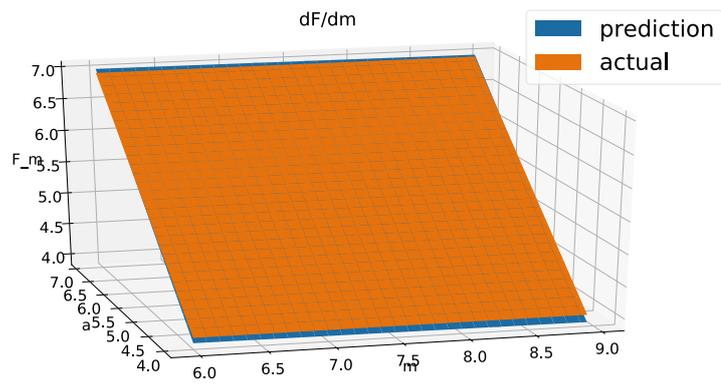

**Figure 4.2.2**

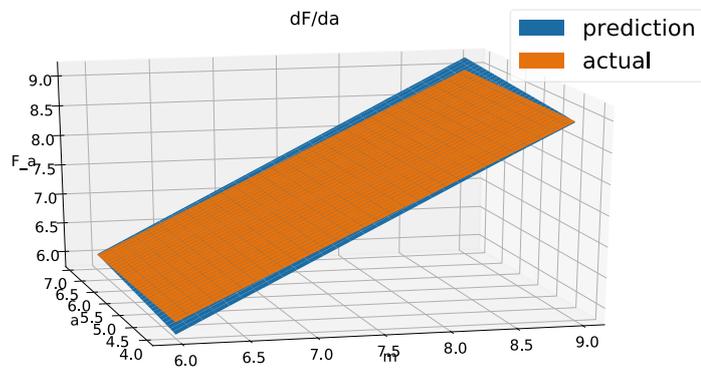

**Figure 4.2.3**

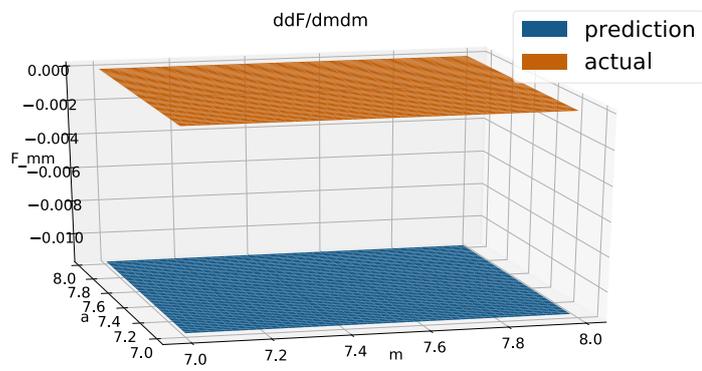

**Figure 4.2.4**

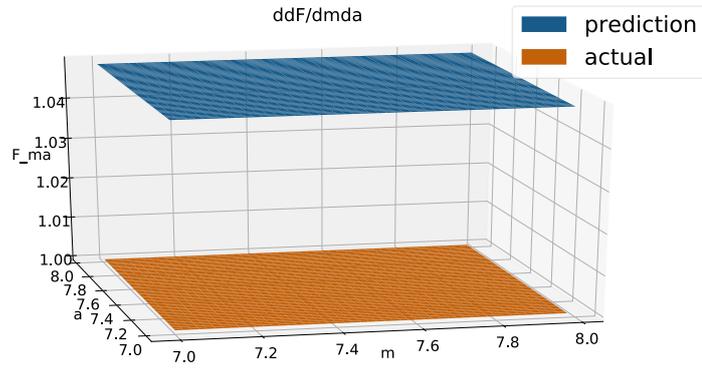

**Figure 4.2.5**

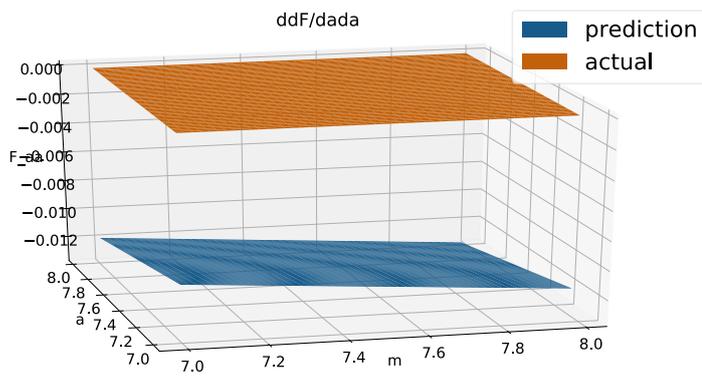

**Figure 4.2.6**

| $F_A(m,a)$ <br> $m,a$ | $f$ | $\dfrac{\partial f}{\partial m}$ | $\dfrac{\partial f}{\partial a}$ | $\dfrac{\partial \partial f}{\partial m \partial m}$ | $\dfrac{\partial \partial f}{\partial m \partial a}$ | $\dfrac{\partial \partial f}{\partial a \partial a}$ |
|---|---|---|---|---|---|---|
| ⟨7,5⟩ | 35 | 5.0 | 6.9 | 0.0 | 1.0 | 0.0 |
| ⟨7,6⟩ | 42 | 6.0 | 7.0 | 0.0 | 1.0 | 0.0 |
| ⟨8,5⟩ | 40 | 4.9 | 8.0 | 0.0 | 1.0 | 0.0 |
| ⟨8,6⟩ | 48 | 6.0 | 8.1 | 0.0 | 1.0 | 0.0 |
| ⟨7.5,5.5⟩ | 41 | 5.5 | 7.5 | 0.0 | 1.0 | 0.0 |

$$P_1(m,a) = 35 + 5.0(m-7) + 6.9(a-5) + 1.0(m-7)(a-5) = ma - 0.1a + 0.5$$

$$P_2(m,a) = 42 + 6.0(m-7) + 7.0(a-6) + 1.0(m-7)(a-6) = ma$$

$$P_3(m, a) = 40 + 4.9(m - 8) + 8.0(a - 5) + 1.0(m - 8)(a - 5) = ma - 0.1m + 0.8$$

$$P_4(m, a) = 48 + 6.0(m - 8) + 8.1(a - 6) + 1.0(m - 8)(a - 6) = ma + 0.1a - 0.6$$

$$P_5(m, a) = 41 + 5.5(m - 7.5) + 7.5(a - 5.5) + 1.0(m - 7.5)(a - 5.5) = ma - 0.25$$

$$\hat{f}(m, a) \approx \frac{1}{5} \sum_{i=1}^{5} P_i(m, a) = ma - 0.02m + 0.09$$

The absolute error upper bound of $\hat{f}(m, a)$ on the sample interval ([5,10], [3,8]): 0.11

The average error of $\hat{f}(m, a)$ on the sample interval: 0.0417

The average error of the original ReLU network on the sample interval: 0.135

## 5. Conclusion

For the first target function, the ANN that directly learns the target function was a bit off in the middle due to a lack of sample data. The result was pretty good. It not only successfully recovered the order of the function, also got a high accuracy on the original interval. The second test result was also ideal. One of the local approximations even perfectly recovers the model. The coefficients of the last two terms in the calculated model were quite low. One may confirm that we can remove those two with further experiments. Additionally, the resulting polynomial has better accuracy than the original ReLU network on the sample interval. Overall, the goals of this paper were successfully met.

## Acknowledgments

I would like to express my gratitude to my supervisor Dr. Nishant Mehta who provided valuable advice in the writing of this paper.

Also, I would like to thank my colleague ZhengYu Zhou who provided a computing device for the experiments.

# Appendix

The following programming tools were used in the numerical experiments

**A.1. Python 3.7.6**

Python 3 is a popular programming language that has a large amount of ML libraries. It is known for its applications in scientific researches and the high readability of the code.

### A.2. NumPy 1.18.1

NumPy is a widely used library for numerical computing. It is also a dependency for a lot of ML libraries such as TensorFlow.

### A.3. TensorFlow 2.1.0

TensorFlow is the most popular ML library in the world today, with great performance and flexibility. It was first released by Google Brain Team in November 2015.

### A.4. Keras 2.3.1

Keras is a library that provides an easy to use tool kit for ML. It can also run on top of other ML libraries like TensorFlow. In the experiments, Keras is used as a high-level API of TensorFlow.

### A.5. Matplotlib 3.2.0

Matplotlib is a very powerful library for drawing figures.

### A.6. Python 3 code of the ANN structure used in the experiments

It includes the function that computes the gradient.

```python
from tensorflow.keras.layers import *
from tensorflow.keras.models import Model, load_model
from numpy import array

class Machine():

    def __init__(self, in_shape = None, model = None, nodes = 1024, lr = 0.01):
        if model:
            self.nnet = load_model(model + ".h5")
        elif in_shape:
            X = Input(in_shape)
            H = Activation('relu')(Dense(nodes)(X))
            Y = Dense(1)(H)
            self.nnet = Model(inputs = X, outputs = Y)
            self.nnet.compile(
                optimizer = Adam(learning_rate = lr),
                loss = 'mean_squared_error'
```

```python
        )
    else:
        raise ValueError("Provide either input shape or model name")

# train
def learn(self, X, Y, ep = 32, bs = 256):
    self.nnet.fit(X, Y, epochs = ep, batch_size = bs)

# compute output
def v(self, data):
    return self.nnet.predict(data)

# compute gradients at given input data points
# only works for one hidden layer for now
def d(self, data):
    gradients = [[0] * len(data) for _ in range(self.nnet.input_shape[1])]
    W = self.nnet.get_weights()
    W1, b, W2 = W[0], W[1], W[2]
    H = array(data) @ W1
    for i in range(len(gradients)):
        for j in range(len(H)):
            for k in range(len(H[j])):
                if H[j][k] > -b[k]:
                    gradients[i][j] += W1[i][k] * W2[k][0]
    return array(gradients)

# save the model as a .h5 file
def save(self, name):
    self.nnet.save(name + ".h5")
```